\documentclass[times,twocolumn,final]{elsarticle}

\usepackage{medima}
\usepackage{framed,multirow}

\usepackage{amssymb}
\usepackage{amsmath}
\DeclareMathOperator*{\argmax}{argmax}
\DeclareMathOperator*{\argmin}{argmin}
\usepackage{latexsym}
\usepackage{subfiles}

\usepackage{url}
\usepackage{xcolor}

\PassOptionsToPackage{hyphens}{url}
\usepackage{url}

\usepackage{breakurl}
\usepackage[breaklinks]{hyperref}
\hypersetup{
    colorlinks=true,
    linkcolor=blue,
    filecolor=magenta,      
    urlcolor=cyan,
}

\usepackage{todonotes}

\definecolor{newcolor}{rgb}{.8,.349,.1}

\begin{document}

\verso{Budd \textit{et~al.}}

\begin{frontmatter}

\title{A Survey on Active Learning and Human-in-the-Loop Deep Learning for Medical Image Analysis\tnoteref{tnote1}}%

\author[1]{Samuel \snm{Budd}\corref{cor1}}
\author[2]{Emma C \snm{Robinson}\fnref{fn1}}
\author[1]{Bernhard \snm{Kainz}\fnref{fn1}}

\address[1]{Department of Computing, Imperial College London, UK}
\address[2]{Department of Imaging Sciences, King's College London, UK}


\begin{abstract}

Fully automatic deep learning has become the state-of-the-art technique for many tasks including image acquisition, analysis and interpretation, and for the extraction of clinically useful information for computer-aided detection, diagnosis, treatment planning, intervention and therapy.
However, 
 the unique challenges posed by medical image analysis suggest that retaining a human end-user in any deep learning enabled system will be beneficial. 
In this review we investigate the role that humans might play in the development and deployment of deep learning enabled diagnostic applications and focus on techniques that will retain a significant input from a human end user. Human-in-the-Loop computing is an area that we see as increasingly important in future research due to the safety-critical nature of working in the medical domain. 
We evaluate four key areas that we consider vital for deep learning in the clinical practice: 
(1) \emph{Active Learning} to choose the best data to annotate for optimal model performance; 
(2) \emph{Interaction with model outputs} - using iterative feedback to steer models to optima for a given prediction and offering meaningful ways to interpret and respond to predictions; 
(3) \emph{Practical considerations} - developing full scale applications and the key considerations that need to be made before deployment; 
(4) \emph{Future Prospective and Unanswered Questions} - knowledge gaps and related research fields that will benefit human-in-the-loop computing as they evolve.
We offer our opinions on the most promising directions of research and how various aspects of each area might be unified towards common goals.


\end{abstract}


\end{frontmatter}

\section{Introduction}
\label{sec1}
Medical imaging is a major pillar of clinical decision making and is an integral part of many patient journeys. Information extracted from medical images is clinically useful in many areas such as computer-aided detection, diagnosis, treatment planning, intervention and therapy. While medical imaging remains a vital component of a myriad of clinical tasks, an increasing shortage of qualified radiologists to interpret complex medical images suggests a clear need for reliable automated methods to alleviate the growing burden on health-care practitioners~\citep{ClinicalReport}.

In parallel, medical imaging sciences are benefiting from the development of novel computational techniques for the analysis of structured data like images. Development of algorithms for image acquisition, analysis and interpretation are driving innovation, particularly in the areas of registration, reconstruction, tracking, segmentation and modelling.

Medical images are inherently difficult to interpret, requiring prior expertise to understand. Bio-medical images can be noisy and contain many modality-specific artefacts, acquired under a wide variety of acquisition conditions with different protocols. Thus, once trained models do not transfer seamlessly from one clinical task or site to another because of an often yawning domain gap~\citep{kamnitsas2017unsupervised,ben2010theory}. Supervised learning methods require extensive relabelling to regain initial performance in different workflows.  

The experience and prior knowledge required to work with such data means that there is often large inter- and intra-observer variability in annotating medical data. This not only raises questions about what constitutes a gold-standard ground truth annotation, but also results in disagreement of what that ground truth truly is. These issues result in a large cost associated with annotating and re-labelling of medical image datasets, as we require numerous expert annotators (oracles) to perform each annotation and to reach a consensus.

In recent years, Deep Learning (DL) has emerged as the state-of-the-art technique for performing many medical image analysis tasks \citep{Tajbakhsh2020EmbracingSegmentation, Tizhoosh2018ArtificialOpportunities., Shen2017DeepAnalysis.,Litjens2017AAnalysis,Suzuki2017OverviewImaging}. 
Developments in the field of computer vision have shown great promise in transferring to medical image analysis, and several techniques have been shown to perform as accurately as human observers \citep{Haenssle2018ManDermatologists,Mar2018ArtificialPromise}. However, uptake of DL methods within the clinical practice has been limited thus far, largely due to the unique challenges of working with complex medical data, regulatory compliance issues and trust in trained models.

We identify three key challenges when developing DL enabled applications for medical image analysis in a clinical setting:

\begin{enumerate}
    \item Lack of Training Data: Supervised DL techniques traditionally rely on a large and even distribution of accurately annotated data points, and while more medical image datasets are becoming available, the time, cost and effort required to annotate such datasets remains significant.
    \item The Final Percent: DL techniques have achieved state-of-the-art performance for medical image analysis tasks, but in safety-critical domains even the smallest of errors can cause catastrophic results downstream. Achieving clinically credible output may require interactive interpretation of predictions (from an oracle) to be useful in practice, i.e users must have the capability to correct and override automated predictions for them to meet any acceptance criteria required.
    \item Transparency and Interpretability: At present, most DL applications are considered to be a 'black-box' where the user has limited meaningful ways of interpreting, understanding or correcting how a model has made its prediction. Credence is a detrimental feature for medical applications as information from a wide variety of sources must be evaluated in order to make clinical decisions. Further indication of how a model has reached a predicted conclusion is needed in order to foster trust for DL enabled systems and allow users to weigh automated predictions appropriately.
\end{enumerate}

There is concerted effort in the medical image analysis research community to apply DL methods to various medical image analysis tasks, and these are showing great promise. 
We refer the reader to a number of reviews of DL in medical imaging 
\citep{Hesamian2019DeepChallenges,Lundervold2019AnMRI,Yamashita2018ConvolutionalRadiology}. These works primarily focus on the development of predictive models for a specific task and demonstrate state-of-the-art performance for that task.

This review aims to give an overview of where humans will remain involved in the development, deployment and practical use of DL systems for medical image analysis. 
We focus on medical image segmentation techniques to explore the role of human end users in DL enabled systems. 

Automating image interpretation tasks like image segmentation suffers from all of the drawbacks incurred by medical image data described above. There are many emerging techniques that seek to alleviate the added complexity of working with medical image data to perform automated segmentation of images. Segmentation seeks to divide an image into semantically meaningful regions (sets of pixels) in order to perform a number of downstream tasks, e.g. biometric measurements. Manually assigning a label to each pixel of an image is a laborious task and as such automated segmentation methods are important in practice. Advances in DL techniques such as Active Learning (AL) and Human-in-the-Loop computing applied to segmentation problems have shown progress in overcoming the key challenges outlined above and these are the studies this review focuses on. We categorise each study based on the nature of human interaction proposed and broadly divide them between which of the three key challenges they address.

Section \ref{sec:al} introduces Active Learning, a branch of Machine Learning (ML) and Human-in-the-Loop Computing that seeks to find the most \textit{informative} samples from an unlabelled distribution to be annotated next. By training on the most informative subset of samples, related work can achieve state-of-the-art performance while reducing the costly annotation burden associated with annotating medical image data.

Section \ref{sec:hitl} evaluates techniques used to refine model predictions in response to user feedback, guiding models towards more accurate per-image predictions. We evaluate techniques that seek to improve interpretability of automated predictions and how models provide feedback on their own outputs to guide users towards better decision making.

Section \ref{sec:prac} evaluates the key practical considerations of developing and deploying Human-in-the-Loop DL enabled systems in practice and outlines the work being done in these areas that addresses the three key challenges identified above. These areas are human focused and assess how human end users might interact with these systems.

In Section \ref{sec:fut} we discuss related areas of ML and DL research that are having an impact on AL and Human-in-the-Loop Computing and are beginning to influence the three key challenges outlined. We offer our opinions on the future directions of Human-in-the-Loop DL research and how many of the techniques evaluated might be combined to work towards common goals. 


\section{Active Learning}\label{sec:al}

\begin{figure*}
    \centering
    \includegraphics[width=\textwidth]{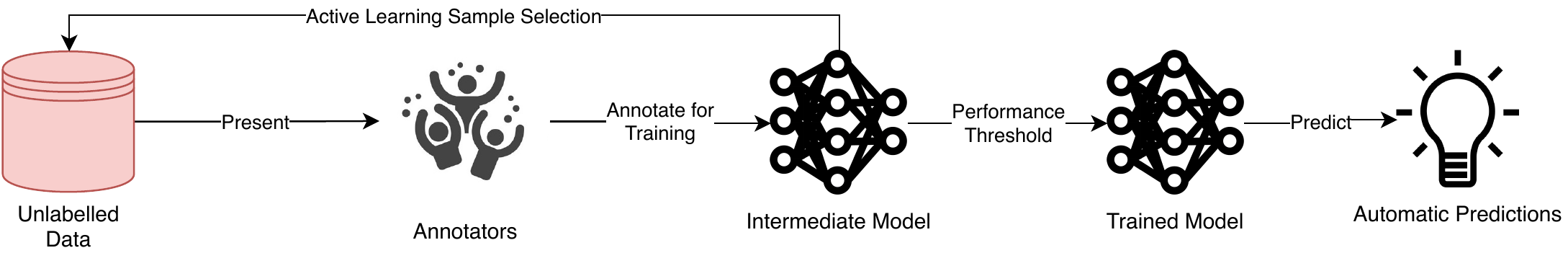}
    \caption{Overview of Active Learning frameworks. }
    \label{fig:al_overview}
\end{figure*}

In this section we assume a scenario in which a large pool of un-annotated data $U$ is available to us, and that we have an oracle (or group of oracles) from which we can request annotations for every un-annotated data point $x_U$ to add to an annotated set $L$. We wish to train some model $f(x|L^*)$ where $L^* \subseteq L$ and consider methods that rely on annotated data to do so. A brute-force solution to this problem would be to ask the oracle(s) to annotate every $x_U$ such that $L^* = L$, but this is rarely a practical or cost-effective solution due to the unique challenges associated with annotating biomedical image data. It is theorised that there is some $L^*$ that achieves equivalent performance to $L$, i.e. $f(x|L^*) \approx f(x|L)$. A model trained on some optimal subset $L^*$ of a dataset might achieve equivalent performance to a model trained on the entire, annotated dataset. Active Learning (AL) is the branch of machine learning that seeks to find this optimal subset $L^*$ given a current model $f'(x|L')$, where $L'$ is an intermediate annotated dataset, and an un-annotated dataset $U$. AL methods aim to iteratively seek the most informative data-points $x^*_i$ for training a model, under the assumption that both the model and the un-annotated dataset will evolve over time, rather than selecting a fixed subset once to be used for training. In a wider context and before the advent of DL, \cite{settles2009active} reviewed this field as a state-of-the-art ML methodology.

A typical AL framework, as outlined in Figure~\ref{fig:al_overview}, consists of a method to evaluate the \textit{informativeness} of each un-annotated data point $x_U$ given $f'(x_U|L')$, tied heavily to the choice of \textit{query type}, after which all chosen data-points are required to be annotated. Once new annotations have been acquired, the AL framework must use the new data to improve the model. This is normally done by either \textit{retraining} the entire model using all available annotated data $L'$, or by \textit{fine-tuning} the network using the most recently annotated data-points $x^*_i$. Using this approach, state-of-the-art performance can be achieved using fewer annotations for several bio-medical image analysis tasks, as shown in the methods discussed in this section, thus widening the annotation bottleneck and reducing the costs associated with developing DL enabled systems from un-annotated data.

\subsection{Query Types}
In every AL framework the first choice to be made is what type of \textit{query} we wish to make using a model and un-annotated dataset. There are currently three main choices available and each lends itself to a particular scenario dependant on what type of un-annotated data we have access to, and what question we wish to ask the oracle(s). 

\textit{Stream-based Selective Sampling} assumes a continuous stream of incoming un-annotated data-points $x_U$ (\cite{NIPS1989_261, Cohn1994ImprovingLearning}). The current model and an \textit{informativeness} measure $I(x_U)$ are used to decide, for each incoming data-point, whether or not to ask the oracle(s) for an annotation (\cite{Dagan1995Committee-BasedClassifiers}). This query type is usually computationally inexpensive but offers limited performance benefits due to the isolated nature of each decision: the wider context of the underlying distribution is not considered, thus balancing exploration and exploitation of the distribution is less well captured than in other query types. Another disadvantage of this query type is calibrating the threshold to use for the chosen informativeness measure such that we do not request annotations for every incoming data-point, and that we do not reject annotations for too many data-points resulting in valuable information being lost. 

\textit{Membership Query Synthesis} assumes that rather than drawing from a real-world distribution of data-points, we instead generate a data-point $x^*_G$ that needs to be annotated (\cite{Angluin1988QueriesLearning}). The generated data-point is what the current model 'believes' will be most informative to itself. This data-point is then annotated by the oracle(s) (\cite{Angluin2001QueriesRevisited}), this can be very efficient in finite domains. This approach may suffer from the same drawbacks as \textit{Stream-based} methods as a model may have no knowledge of unseen areas of the distribution, and thus be unable to request annotations of those areas. Issues can arise where queries can request annotations for data-points that make no sense to a human oracle (\cite{Lang1992QueryUsed}), and are not representative of the actual distribution that is being modelled, stream based and pool based sampling methods were proposed to overcome these issues (\cite{settles2009active}). Nevertheless, recent advances of \textit{Generative Adversarial Networks} (GANs) have shown great promise in  generating data-points that mimic real-world distributions for many different types of data, including biomedical images, that may go someway to addressing the key issue with using query synthesis for complex distributions, which we discuss in Section~\ref{rel:gan}. This query type can be advantageous in scenarios where the distribution to generate is fully understood, or domains in which annotations are acquired autonomously instead of from humans (\cite{King2004FunctionalScientist, King2009TheScience}).

\textit{Pool-based Sampling} assumes a large un-annotated real-world dataset $U$ to draw samples from and seeks to select a batch of \textit{N} samples ${x^*_0, ..., x^*_N}$ from the distribution to request labels for (\cite{Lewis94heterogeneousuncertainty}. \textit{Pool-based} methods usually use the current model to make a prediction on each un-annotated data point to obtain a ranked measure of \textit{informativeness} $I(x_U|f'(x_U|L'))$ for every data-point in the un-annotated set, and select the top \textit{N} samples using this metric to be annotated by the oracle(s). Pool based sampling has been applied to several real world tasks, prior to the advent of deep learning (\cite{Lewis94heterogeneousuncertainty, McCallumEMAL1998, SettlesALS2008, Zhang2002AnAL, HauptmannVideo2006}. These methods can be computationally expensive as every iteration requires a metric evaluation for every data-point in the distribution. However, these methods have shown to be the most promising when combined with DL methods, which inherently rely on a batch-based training scheme. Pool based sampling is used in the majority of methods discussed in the rest of this section unless stated otherwise. While pool-based methods hold advantages over other methods in terms of finding the most informative annotations to acquire, scenarios in which stream based or synthesis based queries are advantageous are also common, such as when memory or processing power is limited for example in mobile or embedded devices (\cite{settles2009active}).

\subsection{Evaluating Informativeness}
In developing an AL framework, once a query type has been selected, the next question to ask is how to measure the informativeness $I(x_U)$ of each of the data-points? Many varying approaches have been taken to quantifying the informativeness of a sample given a model and an underlying distribution. Here we sort these metrics by the level of human interpretability they offer.

Traditionally, AL methods employ hand-designed heuristics to quantify what we as humans believe makes something informative. A variety of model specific metrics seek to quantify what the effect of using a sample for training would have on the model, e.g., the biggest change in model parameters. However, these methods are less prevalent than human designed heuristics due to the computational challenge of applying these to DL models with a large number of parameters. Finally some methods are emerging that are completely agnostic to human interpretability of informativeness and instead seek to learn the best selection policy from available data and previous iterations, as discussed in detail in Section~\ref{sec:lal}.

\subsubsection{Uncertainty}
The main family of informativeness measures falls into calculating uncertainty. It is argued that the more \textit{uncertain} a prediction is, the more information we can gain by including the ground truth for that sample in the training set.

There are several ways of calculating uncertainty from different ML/DL models. When considering DL for segmentation the most simple measure is the sum of lowest class probability for each pixel in a given image segmentation. It is argued that more certain predictions will have high pixel-wise class probabilities, so the lower the sum of the minimum class probability over each pixel in an image, the more certain a prediction is considered to be: 
$$
x^*_{LC} = \argmax_x 1 - P_{\theta}(\hat{y} | x)
$$
where $\hat{y} = \argmax_y P_{\theta}(y | x)$. This is a fairly intuitive way of thinking about uncertainty and offers a means to rank uncertainty of samples within a distribution. We refer to the method above as \textit{least confident} sampling where the samples with the highest uncertainty are selected for labelling~ \citep{settles2009active}. A drawback of \textit{least confident} sampling is that it only considers information about the most probable label, and discards the information about the remaining label distribution. Two alternative methods have been proposed that alleviate this concern. The first, called \textit{margin sampling}~\citep{settles2009active}, can be used in a multi-class setting and considers the first and second most probable labels under the model and calculates the difference between them:
$$
x^*_M = \argmin_x P_{\theta}(\hat{y}_1 | x) - P_{\theta}(\hat{y}_2 | x)
$$
where $\hat{y}_1$ and $\hat{y}_2$ are the first and second most probable labels under the current model, respectively. The intuition here is that the larger the margin is between the two most probable labels, the more confident the model is in assigning that label. The second, more popular approach is to use \textit{entropy} (\cite{ShannonACommunication}) as an uncertainty measure:
$$
x^*_E = \argmax_x - \sum_i P(y_i|x) log P(y_i|x)
$$
where $y_i$ ranges across all possible annotations. Entropy is used to measure the amount of information required to encode a distribution and as such, is often thought of as a measure of uncertainty in
machine learning. For binary classification, all three methods reduce to
querying for the data-point with a class posterior closest to 0.5. The ability of \textit{entropy} to generalise easily to probabilistic multi-class annotations, as well as models for more complex structured data-points has made it the most popular choice for uncertainty based query strategies \cite{SettlesALS2008}.

Using one of the above measures, un-annotated samples are ranked and the most 'uncertain' cases are chosen for the next round of annotation. There have been many recent uses of uncertainty based sampling in AL methods in the DL field and these are discussed next.

\cite{Wang2017Cost-EffectiveClassification} propose the Cost-Effective Active Learning (CEAL) method for deep image classification. The CEAL methods is initialised with a set of unlabelled sample $U$, initially labelled samples $L$, a choice of pool size $K$, a high confidence sample selection threshold $\omega$, a threshold decay rate $dr$, a maximum iteration number $T$ and a fine-tuning interval $t$. After initialisation, CNN weights $W$ are initialised with $L$ and the model is used to make predictions on each data-point in $U$. CEAL explores using each of the three uncertainty methods described above to obtain $K$ uncertain data-points to be manually annotated and added to $D_L$. So far the CEAL method follows very closely the approach outlined in traditional active learning methods as described above, but they introduce an additional training step where the most confident samples (whose \textit{entropy} is less than $\omega$) from $U$ are added to $D_H$. $D_L$ and $D_H$ are then used to fine-tune $W$ for $t$ iterations. CEAL then updates $\omega$ before the \textit{pseudo-labels} from $D_H$ are discarded and each data-point is added back to $U$, while $D_L$ is added to $L$. This process repeats for $T$ iterations. The authors describe this approach of simultaneously learning from manual labels of the most uncertain annotations and predicted labels of the least uncertain annotations as \textit{complementary sampling}. The CEAL method showed that state-of-the-art performance can be achieved using less than 60\% of available data for two non-medical datasets (CACD and Caltech-256) for face recognition and object categorisation.

\cite{Wen2018ComparisonImages.} propose an active learning method that uses uncertainty sampling to support quality control of nucleus segmentation in pathology images. Their work compares the performance improvements achieved through active learning for three different families of algorithms: Support Vector Machines (SVM), Random Forest (RF) and Convolutional Neural Networks (CNN). They show that CNNs achieve the greatest accuracy, requiring significantly fewer iterations to achieve equivalent accuracy to the SVMs and RFs.

Another common method of estimating informativeness is to measure the agreement between multiple models performing the same task. It is argued that more disagreement found between predictions on the same data point implies a higher level of uncertainty. These methods are referred to as \textit{Query by consensus} and are generally applied when \textit{Ensembling} is used to improve performance - i.e, training multiple models to perform the same task under slightly different parameters/settings \cite{settles2009active}. Ensembling methods have shown to measure informativeness well, but at the cost of computational resources - multiple models need to be trained and maintained, and each of these needs to be updated in the presence of newly selected training samples. 

Nevertheless, \cite{BeluchBcaiTheClassification} demonstrate the power of ensembles for active learning and compare to alternatives to ensembling. They specifically compare the performance of acquisition functions and uncertainty estimation methods for active learning with CNNs for image classification tasks and show that ensemble based uncertainties outperform other methods of uncertainty estimation such as 'MC Dropout'. They find that the difference in active learning performance can be explained by a combination of decreased model capacity and lower diversity of MC dropout ensembles. A good performance is demonstrated on a diabetic retinopathy diagnosis task.


\cite{Konyushkova2019GeometrySegmentation} propose an active learning approach that exploits geometric smoothness priors in the image space to aid the segmentation process. They use traditional uncertainty measures to estimate which pixels should be annotated next, and introduce novel criteria for uncertainty in multi-class settings. They exploit geometric uncertainty by estimating the entropy of the probability of supervoxels belonging to a class given the predictions of its neighbours and combine these to encourage selection of uncertain regions in areas of non-smooth transition between classes. They demonstrate state-of-the-art performance on mitochondria segmentation from EM images and on an MRI tumour segmentation task for both binary and multi-class segmentations. They suggest that exploiting geometric properties of images is useful to answer the questions of where to annotate next and by reducing 3D annotations to 2D annotations provide a possible answer to how to annotate the data, and that addressing both jointly can bring additional benefits to the annotation method, however they acknowledge that it would impossible to design bespoke selection strategies this way for every new task at hand.

\cite{Gal2017DeepData} introduce the use of Bayesian CNNs for Active Learning with 'Bayesian Active Learning by Disagreement' or BALD, and show that the use of Bayesian CNNs outperform deterministic CNNs in the context of Active Learning, and exploit this through the use of a new acquisition function that chooses data-points expected to maximise the information gained about the model parameters i.e maximise the mutual information between predictions and model posterior. This approach uses a Bayesian CNN (induced using Dropout during inference~\cite{Gal2016Dropout}), to produce a single prediction using all parameters of the network for each unlabelled data-point, and a set of stochastic predictions for each unlabelled data-point, generated with dropout enabled. The BALD acquisition function is then calculated as the difference between the entropy of the average prediction and average entropy of stochastic predictions. Intuitively this function selects data-points for which the model is uncertain on average, but there exist model parameters that produce disagreeing predicted annotations with high certainty. They demonstrate their approach for skin cancer diagnosis from skin lesion images to show significant performance improvements over uniform sampling using the BALD method for sample selection. While this method has been shown to be particularly effective for AL, when querying batches of data-points, it often results in many very similar, redundant data-points being acquired when used in a greedy fashion, as such BatchBALD was introduced to alleviate this problem \cite{Kirsch2019BatchBALD:Learning}. The BatchBALD approach instead no longer calculates the mutual information between a single sample predictions and model posterior, but instead calculates the mutual information between a batch of samples and the model posterior to jointly score the batch of samples, enabling BatchBALD to more accurately evaluate the joint mutual information and select batches of samples for annotation that result in less redundant data-points being selected together in an acquired batch. This extension is an example of the motivation behind Section \ref{sec:repr} in which we discuss methods that move beyond pure uncertainty based methods and begin to measure diversity among selected samples to reduce redundant annotation.

\subsubsection{Representativeness}\label{sec:repr}
Many AL frameworks extend selection strategies to include some measure of \textit{representativeness} in addition to an uncertainty measure. The intuition behind including a representativeness measure is that methods only concerned with \textit{uncertainty} have the potential to focus only on small regions of the distribution, and that training on samples from the same area of the distribution will introduce redundancy to the selection strategy, or may skew the model towards a particular area of the distribution. The addition of a representativeness measure seeks to encourage selection strategies to sample from different areas of the distribution, and to increase the diversity of samples, thus improving AL performance. A sample with a high representativeness covers the information for many images in the same area of the distribution, so there is less need to include many samples covered by a representative image.

To this end, \cite{Yang2017SuggestiveSegmentation} present Suggestive Annotation, a deep active learning framework for medical image segmentation, which uses an alternative formulation of uncertainty sampling combined with a form of representativeness density weighting. Their method consists of training multiple models that each exclude a portion of the training data, which are used to calculate an ensemble based uncertainty measure. They formulate choosing the most representative example as a generalised version of the maximum set-cover problem (NP Hard) and offer a greedy approach to selecting the most representative images using feature vectors from their models. They demonstrate state-of-the-art performance using 50\% of the available data on the MICCAI Gland segmentation challenge and a lymph node segmentation task.

\cite{Smailagic2018MedAL:Analysis} propose \textit{MedAL}, an active learning framework for medical image segmentation. They propose a sampling method that combines uncertainty, and distance between feature descriptors, to extract the most informative samples from an unlabelled data-set. Once an initial model has been trained, the MedAL method selects data-points to be labelled by first filtering out unlabelled data-points with a predictive entropy below a threshold. From this set the CNN being trained is used to generate feature descriptors for each data-point by taking the output of intermediate layers of the CNN, these feature descriptors are then compared amongst each other using a variety of distance functions (e.g 'Euclidian', 'Russellrao', 'Cosine') in order to find the feature descriptors which are most distant from each other. The data-point with the highest average distance to all other unlabelled data-points (above the entropy threshold) is selected for annotation. In this way, the MedAL acquisition function finds the set of data-points that are both informative to the model, and incur the least redundancy between them by sampling from areas of the input distribution most distant from each other. MedAL method initialises the model in a novel way by leveraging existing computer vision image descriptors to find the images that are most dissimilar to each other and thus cover a larger area of the image distribution to use as the initial training set after annotation. They show good results on three different medical image analysis tasks, achieving the baseline accuracy with less training data than random or pure uncertainty based methods.

\cite{Ozdemir2018ActiveEntropy} propose a Borda-count based combination of an uncertainty and a representativeness measure to select the next batch of samples. Uncertainty is measured as the voxel-wise variance of N predictions using MC dropout in their model. They introduce new representativeness measures such as 'Content Distance', defined as the mean squared error between layer activation responses of a pre-trained classification network. They extend this contribution by encoding representativeness by maximum entropy to optimise network weights using an novel entropy loss function.

\cite{Sourati2018ActiveSegmentation} propose a novel method for ensuring diversity among queried samples by calculating the Fisher Information (FI), for the first time in CNNs. Here, efficient computation is enabled by the gradient computations of propagation to allow FI to be calculated on the large parameter space of CNNs. They demonstrate the performance of their approach on two different flavours of task: a) semi-automatic segmentation of a particular subject (from a different group/different pathology not present in the original training data) where iteratively labelling small numbers of voxels queried by AL achieves accurate segmentation for that subject; and b) using AL to build a model generalisable to all images in a given data-set. They show that in both these scenarios the FI-based AL improves performance after labelling a small percentage of voxels, outperformed random sampling and achieved higher accuracy than entropy based querying.

\subsubsection{Generative Adversarial Networks for Informativeness}\label{rel:gan}

Generative Adversarial Network (GAN) based methods have been applied to several areas of medical imaging such as de-noising, modality transfer, abnormality detection, and for image synthesis, directly applicable to AL scenarios. This offers an alternative (or addition) to the many data augmentation techniques used to expand limited data-sets \cite{Yi2015GenerativeReview} and a DL approach to \textit{Membership Query Synthesis}.

\cite{LastHuman-MachineSegmentation} propose a conditional GAN (cGAN) based method for active learning where they use the discriminator $D$ output as a measure of uncertainty of the proposed segmentations, and use this metric to rank samples from the unlabelled data-set. From this ranking the most uncertain samples are presented to an oracle for segmentation and the least uncertain images are included in the labelled data-set as \textit{pseudo ground truth} labels. They show their method approaches increasing accuracy as the percentage of interactively annotated samples increases - reaching the performance of fully supervised benchmark methods using only 80\% of the labels. This work motivates the use of GAN discriminator scores as a measure of prediction uncertainty.

\cite{Mahapatra2018EfficientNetwork} also use a cGAN to generate chest X-Ray images conditioned on a real image, and using a Bayesian neural network to assess the informativeness of each generated sample, decide whether each generated sample should be used as training data. If so, is used to fine-tune the network. They demonstrate that the approach can achieve comparable performance to training on the fully annotated data, using a dataset where only 33\% of the pixels in the training set are annotated, offering a huge saving of time, effort and costs for annotators.

\cite{Zhao2019DataSegmentation} present an alternative method of data synthesis to GANs through the use of learned transformations. From a single manually segmented image, they leverage other un-annotated images in a SSL like approach to learn a transformation model from the images, and use the model along with the labelled data to synthesise additional annotated samples. Transformations consist of spatial deformations and intensity changes to enable to synthesis of complex effects such as anatomical and image acquisition variations. They train a model in a supervised way for the segmentation of MRI brain images and show state-of-the-art improvements over other one-shot bio-medical image segmentation methods.

The utility of GAN based approaches in AL scenarios goes beyond single-modality image synthesis. Many works have demonstrated the capabilities of GANs to perform cross-modality image synthesis, which directly addresses not only problems of limited training data, but also issues of missing modalities which occur in multi-modal analysis scenarios. Methods by which missing modalities can be generated to fill missing data-points enabling the full suite of AL methods to be applied to multi-modal analysis problems.

\cite{Wang2019DeepMicroscopy} introduce a GAN based method for super-resolution across different microscopy modalities. This work uses GANs to transform diffraction limited input images into super-resolved ones, improving the resolution of wide-field images acquired using low-numerical-aperture objectives to match the resolution acquired using high-numerical-aperture objectives. This work extends this approach to demonstrate cross-modality super-resolution to transform confocal microscopy images to the resolution acquired with a stimulated emission depletion microscope. This approach enables many types of images acquired at lower resolutions to be super-resolved to match those of higher resolutions, enable greater performance of multi-modal image analysis methods in both AL and beyond.

\cite{Wang20183DDose} introduce a GAN based method for the generation of high-quality PET images which usually require a full dose radioactive tracer to obtain. This work enables a low dose tracer to be used to obtain a low-quality PET images, from which a high quality PET image can be generated using a 3D conditional GAN, conditioned on the low-dose image. Additional to this, a 3D c-GANs based progressive refinement scheme is introduced to further improve the quality of estimated images. Through this work the dose of radioactive tracer required to acquire high-quality PET images is greatly reduced, reducing the hazards to patients and enabling low-dose PET images to be used alongside high-dose images in downstream analysis.

\cite{Yu2019Ea-GANs:Synthesis} extend existing GAN based methods for improved cross-modality synthesis of MR images acquired under different scanning parameters. Their work introduces edge-aware generative adversarial networks (Ea-GANs), which specifically integrate edge information reflecting the textural structure of image content to depict the boundaries of different objects in images, which goes beyond methods which focus only on minimising pixel.voxel-wise intensity differences. Using two learning strategies they introduce edge information to a generator-induced Ea-GAN (gEa-GAN) and to a discriminator-induced Ea-GAN (dEa-GAN), incorporating edge information via the generator and both generator and discriminator respectively, so that the edge similarity is also adversarially learned. Their method demonstrates state-of-the-art performance for cross-modal MR synthesis as well as excellent generality to generic image synthesis tasks on facades, maps and cityscapes.

\cite{Pan2020Spatially-ConstrainedNeuroimages} explore the use of GANs to impute missing PET images from corresponding MR images for brain disease identification using a GAN based approach, to avoid discarding data-missing subjects, thus increasing the number of training samples available. A hybrid GAN is used to generate the missing PET images, after which a spatially-constrained Fisher representation network is used to extract statistical descriptors of neuroimages for disease diagnosis. Results on three databases show this method can synthesise reasonable neuroimages and achieve promising results in brain disease identification in comparison to other state-of-the-art methods.

The above works demonstrate the power of using synthetic data conditioned on a very small amount of annotated data to generate new training samples that can be used to train a model to a high accuracy, this is of great value to AL methods where we usually require a initial training set to train a model on before we can employ a data selection policy. These methods also demonstrate the efficient use of labelled data and allow us to generate multiple training samples from a individually annotated image, this may allow the annotated data obtained in AL/Human-in-the-Loop methods to be used more effectively through generating multiple training samples for a single requested annotation, further reducing the annotation effort required to train state-of-the-art models.

\subsubsection{Learning Active Learning}\label{sec:lal}
The majority of methods discussed so far employ hand designed heuristics of informativeness, but some works have emerged that attempt to learn what the most informative samples are through experience of previous sample selection outcomes. This offers a potential way to select samples more efficiently but at the cost of interpretability of the heuristics employed.
Many factors influence the performance and optimality of using hand-crafted heuristics for data selection. \cite{Konyushkova2017LearningData} propose 'Learning Active Learning', where a regression model learns data selection strategies based on experience from previous AL outcomes. Arguing there is no way to foresee the influence of all factors such as class imbalance, label noise, outliers and distribution shape. Instead, their regression model 'adapts' its selection to the problem without explicitly stating specific rules. \cite{Bachman2017LearningLearning} take this idea a step further and propose a model that leverages labelled instances from different but related tasks to learn a selection strategy, while simultaneously adapting its representation of the data and its prediction function.

Reinforcement learning (RL) is a branch of ML that enables an 'agent' to learn in an interactive environment, by trial and error, using feedback from its own actions and experiences, working towards achieving the defined goal of the system. Active Learning has recently been suggested as a potential use-case of RL and several works have begun to explore this area.

\cite{Woodward2017ActiveLearning} propose a one-shot learning method that combines with RL to allow the model to decide, during inference, which examples are worth labelling. A stream of images is presented and a decision is made either to predict the label, or pay to receive the the correct label. Through the choice of RL reward function they are able to achieve higher prediction accuracy than a purely supervised task, or trade prediction accuracy for fewer label requests.

\cite{FangLearningApproach} re-frame the data selection process as a RL problem, and explicitly learn a data selection policy. This is agnostic to the data selection heuristics common in AL frameworks, providing a more general approach, demonstrating improvements in entity recognition, however this is yet to be applied to medical image data.

RL methods offer a different approach to AL and Human-in-the-Loop problems that is well aligned with aiding real-time feedback between a DL enabled application and its end users, however it requires task specific goals that may not be generalisable across different medical image analysis tasks.

\subsection{Fine-tuning vs Retraining}
The final step of each AL framework is to use newly acquired annotations to improve a model. Two main approaches are used to train a model on new annotations. These are retraining the model using all available data including the newly acquired annotations or to fine-tune the model using only new annotations or the new annotations plus a subset from the existing annotations.

\cite{Tajbakhsh2016ConvolutionalTuning} investigate using transfer learning and fine-tuning in several medical image analysis tasks and demonstrate that the use of a pre-trained CNN with fine-tuning outperformed a CNN trained from scratch and that these fine-tuned CNNs were more robust to the size of the training sets. They also showed that neither shallow nor deep tuning was the optimal choice for a particular application and present a layer-wise training scheme that could offer a practical way to reach optimal performance for the chosen task based on the amount of data available. The methods employed in this work perform one-time fine-tuning where a pre-trained model is fine-tuned just once with available training samples, however this does not accommodate an active selection process or continuous fine-tuning.

\cite{ZhouFine-tuning} propose a continuous fine-tuning method that fine-tunes a pre-trained CNN with successively larger datasets and demonstrate that this approach converges faster than repeatedly fine-tuning the pre-trained CNN. They also find that continuously fine-tuning with only newly acquired annotations requires careful meta-parameter adjustments making it less practical across many different tasks.

An alternative approach to retraining from new data that is inspired by the two main approaches described above is to retrain a model using all available data, but using the previous parameters as initialisation, however this approach has not been applied to AL in any works the authors are aware of.

Retraining is computationally more expensive than fine-tuning but it provides a consistent means to evaluate AL framework performance. Fine-tuning is used across a number of different ML areas such as one or few shot learning, and transfer learning and the best approach to this is still an open question and as such is less prevalent in AL frameworks, as fine tuning improves we may see a shift towards its use in AL frameworks. It is important to establish baseline fine-tuning and retraining schemes to effectively compare the DL/AL methods in which they are applied in order to isolate the effects of these schemes from the improvements made in other areas.

\section{The Final Percent: Interactive refinement of model outputs}\label{sec:hitl}

\begin{figure*}
    \centering
    \includegraphics[width=\textwidth]{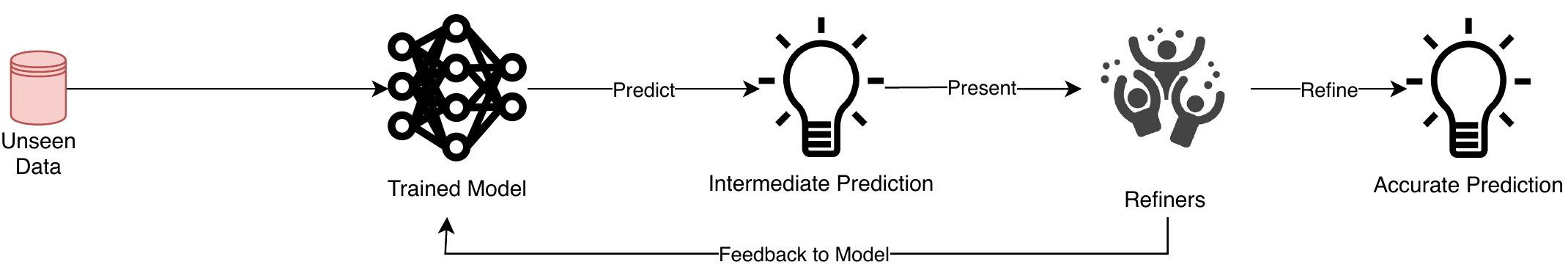}
    \caption{Overview of Refinement frameworks. }
    \label{fig:refinement_overview}
\end{figure*}

So far we have considered the role of humans in annotating data to be used to train a model, but once a model is trained, we still require a human-in-the-loop to interpret model predictions and potentially to refine them to acquire the most accurate results for unseen data, as outlined in Figure~\ref{fig:refinement_overview}. In Human-in-the-loop scenarios, a model makes predictions on unseen input, and subject to acceptance criteria, automated predictions may need manual adjustment to meet those acceptance criteria. Communication of information about the prediction is important to allow acceptance criteria to be met with confidence, and form an understanding of the limitations of automated predictions. This communication is two fold i.e. a user must be able to communicate with the model being used to guide predictions to more accurate results or to correct erroneous predictions, and a model must be able to communicate with the user to provide meaningful interpretation of model predictions, enabling users to take the best course of action when interacting with model outputs and to mitigate human uncertainty. This creates the feedback loop as shown in Figure \ref{fig:refinement_overview}.

\subsection{Interactive Refinement}
If we can develop accurate, robust and interpretable models for medical image applications we still cannot guarantee automated predictions meet acceptance criteria for every unseen data-point presented to a model. The ability to generalise to unseen input is a cornerstone of deep learning applications, but in real world distributions, generalisation is rarely perfect. As such, methods to rectify these discrepancies must be built into applications used for medical image analysis. This iterative refinement must save the end user time and mental effort over performing manual annotation or purely manual correction. Many interactive image segmentation systems have been proposed, and more recently these have built on the advances in deep learning to allow users to refine model outputs and feedback the more accurate results to the model for improvement.

\cite{Amrehn2017UI-Net:Model} introduced UI-Net, that builds on the popular U-Net architecture for medical image segmentation \cite{RonnebergerU-Net:Segmentation}. The UI-Net is trained with an \textit{active user model}, and allows for users to interact with proposed segmentations by providing \textit{scribbles} over the image to indicate areas that should be included or not, the network is trained using simulated user interactions and as such responds to iterative user scribbles to refine a segmentation towards a more accurate result.

Conditional Random fields have been used in various tasks to encourage segmentation homogeneity. \cite{ZhengConditionalNetworks} propose CRF-CNN, a recurrent neural network which has the desirable properties of both CNNs and CRFs. \cite{WangDeepIGeoS:Segmentation} propose DeepIGeoS, an interactive geodesic framework for medical image segmentation. This framework uses two CNNs, the first performs an initial automatic segmentation, and the second takes the initial segmentation as well as user interactions with the initial segmentation to provide a refined result. They combine user interactions with CNNs through geodesic distance transforms \cite{Criminisi2008GeoS:Segmentation}, and these user interactions are integrated as hard constraints into a Conditional Random Field, inspired by \cite{ZhengConditionalNetworks}. They call their two networks P-Net (initial segmentation) and R-Net (for refinement). They demonstrate superior results for segmentation of the placenta from 2D fetal MRI and brain tumors from 3D FLAIR images when compared to fully automatic CNNs. These segmentation results were also obtained in roughly a third of the time taken to perform the same segmentation with traditional interactive methods such as GeoS or ITK-SNAP.

Graph Cuts have also been used in segmentation to incorporate user interaction - a user provides \textit{seed points} to the algorithm (e.g. mark some pixel as foreground, and another as background) and from this the segmentation is calculated. \cite{WangInteractiveFine-tuning} propose BIFSeg, an interactive segmentation framework inspired by graph cuts. Their work introduces a deep learning framework for interactive segmentation by combining CNNs with a bounding box and scribble based segmentation pipeline. The user provides a bounding box around the area which they are interested in segmenting, this is then fed into their CNN to produce an initial segmentation prediction, the user can then provide scribbles to mark areas of the image as mis-classified - these user inputs are then weighted heavily in the calculation of the refined segmentation using their graph cut based algorithm.

\cite{BredellIterativeNetworks} propose an alternative to BIFSeg in which two networks are trained, one to perform an initial segmentation (they use a CNN but this initial segmentation could be performed with any existing algorithm) and a second network they call interCNN that takes as input the image, some user scribbles and the initial segmentation prediction and outputs a refined segmentation, they show that with several iterations over multiple user inputs the quality of the segmentations improve over the initial segmentation and achieve state-of-the-art performance in comparison to other interactive methods.

The methods discussed above have so far been concerned with producing segmentations for individual images or slices, however many segmentation tasks seek to extract the 3D shape/surface of a particular region of interest (ROI). \cite{Kurzendorfer2017Rapid} propose a dual method for producing segmentations in 3D based on a Smart-brush 2D segmentation that the user guides towards a good 2D segmentation, and after a few slices are segmented this is transformed to a 3D surface shape using Hermite radial basis functions, achieving high accuracy. While this method does not use deep learning it is a strong example of the ways in which interactive segmentation can be used to generate high quality training data for use in deep learning applications - their approach is general and can produce segmentations for a large number of tasks. There is potential to incorporate deep learning into their pipeline to improve results and accelerate the interactive annotation process.

\cite{jang2019interactive} propose an interactive segmentation scheme that generalises to any previously trained segmentation model, which accepts user annotations about a target object and the background. User annotations are converted into interaction maps by measuring the distance of each pixel to the annotated landmarks, after which the forward pass outputs an initial segmentation. The user annotated points can be mis-segmented in the initial segmentation so they propose BRS (back-propogating refinement scheme) that corrects the mis-labelled pixels. They demonstrate that their algorithm outperforms conventional approaches on several datasets and that BRS can generalise to medical image segmentation tasks by transforming existing CNNs into user-interactive versions.

\cite{Liao2020IterativelyRefinedI3} propose modelling the dynamics of iterative interactive refinement as a Markov Decision Process (MDP) and solve this with multi-agent RL. Treating each voxel as an agent with a shared voxel-level behaviour strategy they make voxel-wise prediction tractable in this way. The multi-agent method successfully captures the dependencies among voxels for segmentation tasks, and by passing prediction uncertainty of previous segmentations through the state space can derive more precise and finer segmentations. Using this method they significantly outperform existing state-of-the-art methods with fewer interactions and a faster convergence.

In this section we focus on applications concerned with iteratively refining a segmentation towards a desired quality of output. In the scenarios above this is performed on an un-seen image provided by the end user, but there is no reason the same approach could not be taken to generate iteratively more accurate annotations to be used in training, e.g., using active learning to select which samples to annotate next, and iteratively refining the prediction made by the current model until a sufficiently accurate annotation is curated. This has the potential to accelerate annotation for training without any additional implementation overhead. Much work done in AL ignores the role of the oracle and merely assumes we can acquire an accurate label when we need it, but in practice this presents a more significant challenge. We foresee AL and HITL computing become more tightly coupled as AL research improves it's consideration for the oracle providing the annotations.

It is fairly intuitive how a user might refine segmentations of medical images, but this is not the case for other medical image analysis tasks. Refinements of predictions on clinical tasks involving classification and regression have seen less development than those in segmentation and remains an open area of research. The following works have taken steps towards addressing interactive refinement strategies for classification and regression tasks.

\cite{Lian2020HierarchicalMRI} explore the use of CNN methods for automated diagnosis of Alzheimer's disease and identify that many state-of-the-art methods rely on the pre-determination of informative locations in structural MRI (sMRI). This stage of discriminative localisation is isolated from the latter stages of feature extraction and classifier construction. Their work proposes a hierarchical fully convolutional CNN (H-FCN) to automatically identify discriminative local patches and regions in whole brain sMRI, from which multi-scale feature representations can be jointly learned and fused to construct classification models. This work enables interactive refinement of patch choice and classifier construction which, if intervened on by human end users could guide the network towards more discriminative regions of interest and thus more effective classifiers.

Similarly, \cite{Liu2018Landmark-basedDiagnosis} introduce a landmark-based deep multi-instance learning (LDMIL) framework for brain disease diagnosis. Firstly, by adopting a data-driven approach to discover disease related anatomical landmarks in brain MR images, along with nearby image patches. Secondly the framework learns an end-to-end MR image classifier for capturing local structural information in the selected landmark patches, and global structure information derived from all detected landmarks. By splitting the steps of landmark detection and classifier construction, a human-in-the-loop can be introduced to intervene on selected landmarks and to guide the network towards maximally informative image regions. Thus, the resulting classifier can be refined via updating which regions of the image are used as input.

\subsection{Interactive Interpretation}
In the previous section we discussed methods by which the user of a human-in-the-loop system might communicate with a predictive model, in this section we consider methods by which a model might communicate with the user, thus completing the feedback loop in Figure \ref{fig:refinement_overview}. 'Interpretation' can mean many different things depending on the context, so here we focus on interpretation of model outputs with the goal of appropriately weighting automated predictions in downstream analysis (e.g uncertainty of predictions) and to enable users to make the most informed corrections or manual adjustments to model predictions (e.g 'Attention Gating'\cite{OktayAttentionPancreas}).

While DL methods have become a standard state-of-the-art approach for many medical image analysis tasks, they largely remain black-box methods where the end user has limited meaningful ways of interpreting model predictions. This feature of DL methods is a significant hurdle in the deployment of DL enabled applications to safety-critical domains such as medical image analysis. We want models to be highly accurate and robust, but also explainable and interpretable. This interpretability is vital to mitigate human uncertainty and foster trust in using automated predictions in downstream tasks with real-world consequences.

Recent EU law\footnote{Regulation (EU) 2016/679 on the protection of natural persons with regard to the processing of personal data and on the free movement of such data, and repealing Directive 95/46/EC (General Data Protection Regulation) [2016] OJ L119/1} has led to the 'right for explanation', whereby any subject has the right to have automated decisions that have been made about them explained. This even further highlights the need for transparent algorithms which we can reason about [\cite{Goodman2016EuropeanExplanationquot}, \cite{Edwards2017EnslavingDecisionss}, \cite{Edwards2017SlaveFor}].

It is important for users to understand how a certain decision has been made by the model, as even the most accurate and robust models aren't infallible, and false or uncertain predictions must be identified so that trust in the model can be fostered and predictions are appropriately weighted in the clinical decision making process. It is vital the end user, regulators and auditors all have the ability to contextualise automated decisions produced by DL models. Here we outline some different methods for providing interpretable ways of reasoning about DL models and their predictions.

Typically DL methods can provide statistical metrics on the uncertainty of a model output, many of the uncertainty measures discussed in Section \ref{sec:al} are also used to aid in intepretability. While uncertainty measures are important, these are not sufficient to foster complete trust in DL model, the model should provide human-understandable justifications for its output that allow insights to be drawn elucidating the inner workings of a model. \cite{Chakraborty2017InterpretabilityResults} discuss many of the core concerns surrounding model intepretability and highlight various works that have demonstrated sophisticated methods of making a DL model interpretable across the DL field. Here we evaluate some of the works that have been applied to medical image segmentation and refer the reader to [\cite{2018UnderstandingApplications}, \cite{Holzinger2017TowardsPathology}] for further reading on interpretability in the rest of the medical imaging domain.

\cite{OktayAttentionPancreas} and \cite{Schlemper2019AttentionImages} introduce 'Attention Gating' to guide networks towards giving more 'attention' to certain image areas, in a visually interpretable way - potentially aiding in the subsequent refinement of annotations. Attention Gates are introduced into the popular U-Net architecture (\cite{RonnebergerU-Net:Segmentation}), where information extracted from coarse scale layers is used in gating to disambiguate irrelevant and noisy responses in skip connections, prior to concatenation, to merge only relevant layer activations. This approach eliminates the need for applying external object localisation models in image segmentation and regression tasks. Coefficients of Attention Gate layers indicate where in an image feature activations will be allowed to propagate through to final predictions, providing users with a visual representation of the areas of an image that a model has weighted highly in making predictions.

In \cite{Budd2019ConfidentSonographers} we propose a visual method for interpreting automated head circumference measurements from ultrasound images, using MC Dropout at test-time to acquire N head segmentations to calculate an upper and lower bound on the head circumference measurement in real-time. These bounds were displayed over the image to guide the sonographer towards views in which the model predicts with the most confidence. This upper lower bound is presented as a measure of model compliance of the unseen image rather than uncertainty. Finally, variance heuristics are proposed to quantify the confidence of a prediction in order to either accept or reject head circumference measurements, and it is shown these can improve overall performance measures once 'rejected' images are removed.

\cite{Milletari2019StraightUltrasound} propose the application of RL to ultrasound care, guiding a potentially inexperienced user to the correct sonic window and enabling them to obtain clinically relevant images of the anatomy of interest. This human-in-the-loop application is an example of the novel applications possible when combining DL/RL with real-time systems enabling users to respond to model feedback to acquire the most accurate information available.

\cite{Wang2019AleatoricNetworks} propose using test-time augmentation to acquire a measure of aleatoric (image-based) uncertainty and compare their method with epistemic (model) uncertainty measures and show that their method provides a better uncertainty estimation than a test-time dropout based model uncertainty alone and reduces overconfident incorrect predictions.

\cite{Jungo2019AssessingSegmentation} evaluate several different voxel-wise uncertainty estimation methods applied to medical image segmentation with respect to their reliability and limitations and show that current uncertainty estimation methods perform similarly. Their results show that while uncertainty estimates may be well calibrated at the dataset level (capturing epistemic uncertainty), they tend to be mis-calibrated at a subject-level (aleatoric uncertainty). This compromises the reliability of these uncertainty estimates and highlights the need to develop subject-wise uncertainty estimates. They show auxiliary networks to be a valid alternative to common uncertainty methods as they can be applied to any previously trained segmentation model.

Developing transparent systems will enable faster uptake in clinical practice and including humans within the deep learning clinical pipelines will ease the period of transition between current best practices and the breadth of possible enhancements that deep learning has to offer. 

We suggest that ongoing work in improving interpretability of DL models will also have a positive impact on AL, as the majority of methods to improve intepretability are centred on providing uncertainty measures for a models prediction, these same uncertainty measures can be used for AL selection strategies in place of existing uncertainty measures that are currently employed. As intepretability and uncertainty measures improve we expect to see a similar improvement of AL frameworks as they incorporate the most promising uncertainty measures.

The methods discussed in Section \ref{sec:hitl} remain open areas of research interest with great implications for the progress of AL development and greater uptake of DL and HITL methods in clinical practice. The study of interaction between users and models is of growing importance and is having a significant impact on the efficacy of Deep Active Learning systems and their deployment to real-world applications, especially in clinical scenarios (\cite{Beede2020ARetinopathy, Amrehn2019ASystems}). The wider study of interpretability in ML and the study of Human Computer Interaction may seem distinct and diverging, however we expect to see these two research fields converge through Active Learning as the feedback loop between human users and machine models becomes of increasing importance.

\section{Practical Considerations}\label{sec:prac}

\begin{figure*}
    \centering
    \includegraphics[width=\textwidth]{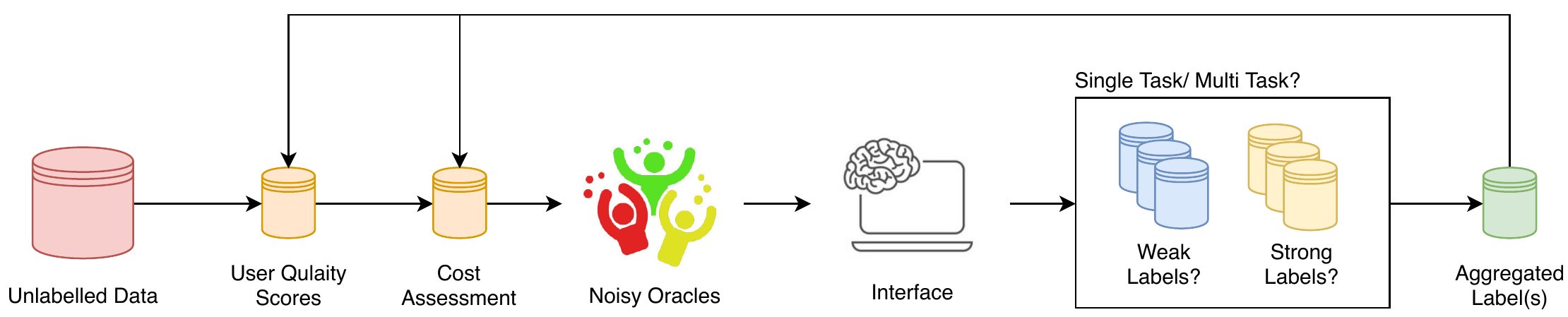}
    \caption{Overview of practical considerations}
    \label{fig:prac_overview}
\end{figure*}

We have so far discussed the core body of work behind AL, model interpretation and prediction refinement, and while the works discussed above go a long way in covering the majority of research being done, there are several practical considerations for developing and deploying DL enabled applications that we must consider. In this section we outline the main practical research areas that are impacting DL enabled application development pipelines and suggest where we might look next.

\subsection{Noisy Oracles}
Gold-standard annotations for medical image data are acquired by aggregating annotations from multiple expert oracles, but as previously discussed, this is rarely feasible to obtain for large complex datasets due to the expertise required to perform such annotations. Here we ask what effect on performance we might incur if we acquire labels from oracles without domain expertise, and what techniques can we use to mitigate the suspected degradation of annotation quality when using non-expert oracles, to avoid any potential loss in accuracy.

\cite{Zhang2015ActiveLabelers} propose active learning method that assume data will be annotated by a crowd of non-expert or 'weak' annotators, and offer approaches to mitigate the introduction of bad labels into the data set. They simultaneously learn about the quality of individual annotators so that the most informative examples can be labelled by the strongest annotators.

\cite{Li2018Crowds} propose methods for crowd-sourced learning in two scenarios. Firstly, they aim at inferring instances ground truth given the crowd's annotations by modelling the crowd's expertise and label correlations from two different perspectives: firstly they model expertise based on individual labels, based on the idea that labeller's annotations for similar instances should be similar, and secondly through modelling the crowd's expertise to distinguish the relevance between label pairs. They extend their approach to the active paradigm and offer criteria for instance, label and labeller selected in tandem to minimise annotation cost.

\cite{CheplyginaEarlyCT} explore using Amazon's MTurk to gather annotations of airways in CT images. Results showed that the novice oracles were able to interpret the images, but that instructions provided were too complex, leading to many unusable annotations. Once the bad annotations were removed, the annotations did show medium to high correlation with expert annotations, especially if annotations were aggregated.

\cite{RodriguesDeepCrowds} describe an approach to assess the reliability of annotators in a crowd, and a crowd layer used to train deep models from noisy labels from multiple annotators, internally capturing the reliability and biases of different annotators to achieve state-of-the-art results for several crowd-sourced data-set tasks.

We can see that by using a learned model of oracle annotation quality we can mitigate the effects of low quality annotations and present the most challenging cases to most capable oracles. By providing clear instructions we can lower the barriers for non-expert oracles to perform accurate annotation, but this is not generalisable and would be required for every new annotation task we wish to perform.

\subsection{Weakly Supervised Learning}
Most segmentation tasks require pixel-wise annotations, but these are not the only type of annotation we can give an image. Segmentation can be performed with 'weak' annotations, which include image level labels e.g. modality, organs present etc. and annotations such as bounding boxes, ellipses or scribbles. It is argued that using 'weaker' annotation formulations can make the task easier for the human oracle, leading to more accurate annotations. 'Weak' annotations have been shown to perform well in several segmentation tasks,  \cite{RajchlDeepCut:Networks} demonstrate obtaining pixel-wise segmentations given a data-set of images with 'weak' bounding box annotations. They propose DeepCut, an architecture that combines a CNN with an iterative dense CRF formulation to achieve good accuracy while greatly reducing annotation effort required. In a later study, \cite{RajchlEmployingProblems} examine the impact of expertise required for different 'weak' annotation types on the accuracy of liver segmentations. The results showed a decrease in accuracy with less expertise, as expected, across all annotation types. Despite this, segmentation accuracy was comparable to state-of-the-art performance when using a weakly labelled atlas for outlier correction. The robust performance of their approach suggests 'weak' annotations from non-expert crowds could be used to obtain accurate segmentations on many different tasks, however their use of an atlas makes this approach less generalisable than is desired.

In \cite{RajchlLearningSupervision} they examine using super pixels to accelerate the annotation process. This approach uses a pre-processing step to acquire a super-pixel segmentation of each image, non-experts are then used to perform the annotation by selecting which super-pixels are part of the target region. Results showed that the approach largely reduces the annotation load on users. Non-expert annotation of 5000 slices was completed in under an hour by 12 annotators, compared to an expert taking three working days to establish the same with an advanced interface. The non-expert interface is web-based demonstrating the potential of distributed annotation collection/crowd-sourcing. An encouraging aspect of this paper is that the results showed high performance on the segmentation task in question compared with expert annotation performance, but may not be suitable for all medical image analysis tasks.

It has been shown that we can develop high performing models using weakly annotated data, and as weak annotations requires less expertise to perform, they can be acquired faster and from a non-expert crowd with a smaller loss in accuracy than gold-standard annotations. This is very promising for future research as datasets of weakly annotated data might be much easier and more cost-effective to curate.

\subsection{Multi-task learning}
Many works aim to train models or acquire training data for several tasks at once, it is argued that this can save on cost as complementary information may result in higher performance over multiple different tasks \citep{Moeskops2016DeepModalities}. 
\cite{Wang2019Mixed-SupervisedSegmentation} propose a dual network for joint segmentation and detection task for lung nodule segmentation and cochlea segmentation from CT images, where only a part of the data is densely annotated and the rest is weakly labelled by bounding boxes, using this they show that their architecture out-performs several baselines. At present this work only handles the case for two different label types but they propose extending the framework for a true multi-task scenario.

This is a promising area but, as of yet, it has not been incorporated into an active learning setting. As such, it may be elucidating to analyse the differences in samples chosen by different AL methods when the model is being training for multiple tasks simultaneously. However, \cite{LowellPracticalLearning} raise concerns over the transferability of actively acquired datasets to future models due to the inherent coupling between active learning selection strategies and the model being trained, and show that training a successor model on the actively acquired dataset can often result in worse performance than from random sampling. They suggest that, as datasets begin to outlive the models trained on them, there is a concern for the efficacy of active learning, since the acquired dataset may be disadvantageous for training subsequent models. An exploration of how actively acquired datasets perform on multiple models may be required to explain the effects of an actively acquired dataset coupled with one model on the performance of related models.

\subsection{Annotation Interface}
So far the majority of Human-in-the-loop methods assume a significant level of interaction from an oracle to annotate data and model predictions, but few consider the nature of the interface with which an oracle might interact with these images. The nature of medical images require special attention when proposing distributed online platforms to perform such annotations. While the majority of techniques discussed so far have used pre-existing data labels in place of newly acquired labels to demonstrate their performance, it is important to consider the effects of accuracy of annotation that the actual interface might incur. 

\cite{Nalisnik2015AnImages.} propose a framework for the online classification of Whole-slide images (WSIs) of tissues. Their interface enables users to rapidly build classifiers using an active learning process that minimises labelling efforts and demonstrates the effectiveness of their solution for the quantification of glioma brain tumours.

\cite{Khosravan2017Gaze2Segment:Segmentation} propose a novel interface for the segmentation of images that tracks the users gaze to initiate seed points for the segmentation of the object of interest as the only means of interaction with the image, achieving high segmentation performance. \cite{Stember2019EyeNetworks} extend this idea and compare using eye tracking generated training samples to traditional hand annotated training samples for training a DL model. They show that almost equivalent performance was achieved using annotation generated through eye tracking, and suggest that this approach might be applicable to rapidly generate training data. They acknowledge that there is still improvements to be made integrate eye tracking into typical clinical radiology work flow in a faster, more natural and less distracting way.

\cite{Tinati2017AnProject} evaluate the player motivations behind EyeWire, an online game that asks a crowd of players to help segment neurons in a mouse brain. The gamification of this task has seen over 500,000 players sign up and the segmentations acquired have gone onto be used in several research works [\cite{Kim2014SpacetimeRetina}]. One of the most exciting things about gamification is that when surveyed, users were motivated most by making a scientific contribution rather than any potential monetary reward. However this is very specialised towards this particular task and would be difficult to apply across other types of medical image analysis tasks.

There are many different approaches to developing annotation interfaces and the ones we consider above are just a few that have been applied to medical image analysis. As development increases we expect to see more online tools being used for medical image analysis and the chosen format of the interface will play a large part in the usability and overall success of these applications.

\subsection{Variable Learning Costs}
When acquiring training data from various types of oracle it is worth considering the relative cost associated with querying a particular oracle type for that annotation. We may wish to acquire more accurate labels from an expert oracle, but this is likely more expensive to obtain than from a non-expert oracle. The trade off, of course, being accuracy of the obtained label - less expertise of the oracle will likely result in a lower quality of annotation. Several methods have been proposed to model this and allow developers to trade off between cost and overall accuracy of acquired annotations.

\cite{Kuo2018Cost-SensitiveDetection} propose a cost-sensitive active learning approach for intracranial haemorrhage detection. Since annotation time may vary significantly across examples, they model the annotation time and optimize the return on investment. They show their approach selects a diverse and meaningful set of samples to be annotated, relative to a uniform cost model, which mostly selects samples with massive bleeds which are time consuming to annotate.

\cite{ShahAnnotation-costNetworks} propose a budget based cost minimisation framework in a mixed-supervision setting (strong and weak annotations) via dense segmentation, bounding boxes, and landmarks. Their framework uses an uncertainty and a representativeness ranking strategy to select samples to be annotated next. They demonstrate state-of-the-art performance at a significantly reduced training budget, highlighting the important role of choice of annotation type on the costs of acquiring training data.

The above works each show an improved consideration for the economic burden that is incurred when curating training data. A valuable research direction would be to assess the effects of oracle expertise level, annotation type and image annotation cost in a unified framework as these three factors are very much linked and may have a profound influence over each other.


\section{Future Prospective and Unanswered Questions}\label{sec:fut}

In Sections \ref{sec:al} \& \ref{sec:hitl} we discuss methods through which a user might gather training data to build a model, use their model to predict on new data and receive feedback to iteratively refine the model output towards a more accurate result. Each of these techniques assume some human end user will be present to interact with the system at the point of initial annotation, interpretation and refinement. Each of these areas seeks to achieve a shared goal of achieving the highest performing model from as little annotated data as possible - with a means to weigh conclusions of models predictions appropriately. 

AL is not the only area of research that aim to learn from limited data. Semi-supervised learning, and Transfer Learning both make significant contributions to extracting the most value from limited labelled data.

In the presence of large data-sets, but the absence of labels, unsupervised and semi-supervised approaches offer a means by which information can be extracted without requiring labels for all the data-points. This could potentially have a massive impact on the medical image analysis field where this is often the case.

In a semi-supervised learning (SSL) scenario we may have some labelled data, but this is often very limited. We do however have a large set of un-annotated instances (much like in active learning) to draw information from, the goal being to improve a model (trained only on the labelled instances) using the un-labelled instances. From this we derive two distinct goals: a) predicting labels for future data (inductive SSL) and b) predicting labels for the available un-annotated data (transductive SSL) (\cite{CheplyginaNot-so-supervised:Analysis, 9093608}). SSL methods provide a powerful way of extracting useful information from un-annotated image data and we believe that progress in this area will be beneficial to AL systems that desire a more accurate model for initialisation to guide data selection strategies.

Transfer Learning (TL) is a branch of DL that aim to use pre-trained networks as a starting point for new applications. Given a pre-trained network trained for a particular task, it has been shown that this network can be 'fine-tuned' towards a target task from limited training data. We refer the reader to \cite{Morid2021AImageNet,Raghu2019TransfusionUT,CheplyginaNot-so-supervised:Analysis} for a more general overview of transfer learning in medical imaging, and focus on the use of TL in AL scenarios in the following.  \cite{Tajbakhsh2016ConvolutionalTuning} demonstrated the applicability of TL for a variety of medical image analysis tasks, and show, despite the large differences between natural images and medical images, CNNs pre-trained on natural images and fine-tuned on medical images can perform better than medical CNNs trained from scratch. This performance boost was greater where fewer target task training examples were available. Many of the methods discussed so far start with a network pre-trained on natural image data.

\cite{ZhouAFTEfforts} propose AFT*, a platform that combines AL and TL to reduce annotation efforts, which aims at solving several problems within AL. AFT* starts with a completely empty labelled data-set, requiring no seed samples. A pre-trained CNN is used to seek 'worthy' samples for annotation and to gradually enhance the CNN via continuous fine-tuning. A number of steps are taken to minimise the risk of catastrophic forgetting. Their previous work \cite{ZhouFine-tuning} applies a similar but less featureful approach to several medical image analysis tasks to demonstrate equivalent performance can be reached with a heavily reduced training data-set. They then use these tasks to evaluate several patterns of prediction that the network exhibits and how these relate to the choice of AL selection criteria.

\cite{Zhou2018IntegratingInterpretation} have gone onto to use their AFT framework for annotation of CIMT videos, a clinical technique for characterisation of Cardiovascular disease. Their extension into the video domain presents its own unique challenges and thus they propose a new concept of an Annotation Unit - reducing annotating a CIMT video to just 6 user mouse clicks, and by combining this with their AFT framework reduce annotation cost by 80\% relative to training from scratch and by 50\% relative to random selection of new samples to be annotated (and used for fine-tuning).

\cite{Kushibar2019SupervisedInteraction} use TL for supervised domain adaptation for sub-cortical brain structure segmentation with minimal user interaction. They significantly reduce the number of training images from different MRI imaging domains by leveraging a pre-trained network and improve training speed by reducing the number of trainable parameters in the CNN. They show their method achieves similar results to their baseline while using a remarkably small amount of images from the target domain and show that using even one image from the target domain was enough to outperform their baseline.

The above methods and more discussed in this review demonstrate the applicability of TL to reducing the number of annotated sample required to train a model on a new task from limited training data. By using pre-trained networks trained on annotated natural image data (there is an abundance) we can boost model performance and further reduce the annotation effort required to achieve state-of-the-art performance. 

A related sub-field of TL worth exploring is domain adaptation (DA). Many DL techniques used in medical image analysis suffer from the domain shift problem caused by different distributions between source data and target data, often due to medical images being acquired on a variety of different scanners, scanning parameters and subject cohorts etc. DA has been proposed as a special type of transfer learning in which the domain feature space and tasks remain the same while marginal distributions of the source and target domains are different. We refer the reader to \cite{Guan2021DomainSurvey,Choudhary2020AdvancingAdaptation} for an overview of DA methods used for medical image analysis, and hope to see greater application of DA methods in AL scenarios in the future.

In many of scenarios described in this review, models continuously receive new annotations to be used for training, and in theory we could continue to retrain or fine-tune a model indefinitely, but is this practical and cost effective? It is important to quantify the long term effects of training a model with new data to assess how the model changes over time and whether or not performance has improved, or worse, declined. Learning from continuous streams of data has proven more difficult than anticipated, often resulting in 'catastrophic forgetting' or 'interference' \cite{ParisiContinualReview}. We face the \textit{stability-plasticity-dilemma}. Avoiding catastrophic forgetting in neural networks when learning from continuous streams of data can be broadly divided among three conceptual strategies: a) Retraining the the whole network while regularising (to prevent forgetting of previously learned tasks). b) selectively train the network and expand it if needed to represent new tasks, and c) retaining previous experience to use memory replay to learn in the absence of new input. We refer the reader to \cite{ParisiContinualReview} for a more detailed overview of these approaches. 

\cite{BawejaTowardsImaging} investigate continual learning of two MRI segmentation tasks with neural networks for countering catastrophic forgetting of the first task when a new one is learned. They investigate elastic weight consolidation, a method based on Fisher information to sequentially learn segmentation of normal brain structures and then segmentation of white matter lesions and demonstrate this method reduces catastrophic forgetting, but acknowledge there is a large room for improvement for the challenging setting of continual learning.

It is important to quantify the performance and robustness of a model at every stage of its lifespan. One way to consider stopping could evaluate when the cost of continued training outweighs the cost of errors made by the current model. An existing measure that attempts to quantify the economical value of medical intervention is the Quality-adjusted Life year (QALY), where one QALY equates to one year of healthy life \cite{JudgingNICE}. Could this metric be incorporated into models? At present we cannot quantify the cost of errors made by DL medical imaging applications but doing so could lead to a deeper understanding of how accurate a DL model really ought to be.

As models are trained on more of the end user's own data, will this cause the network to perform better on data from that user's system despite performing worse on data the model was initially trained on? Catastrophic forgetting suggests this will be the case, but is this a bad thing? It may be beneficial for models to gradually bias themselves towards high performance for the end user's own data, even if this results in the model becoming less transferable to other data. \cite{farquhar2021on} explore the role of bias in AL methods. Bias is introduced because the training data no longer follows the population distribution in AL. The authors providing a general method by which unbiased AL estimators may be constructed using novel corrective weights to remove bias. Further to this, an explanation of the empirical successes of existing AL methods which ignore this bias is provided. It is shown that bias introduced by AL methods can be actively helpful when training overparameterized models like neural networks with relatively little data. This further motivates future work to better understand when the bias introduced by AL could have a positive influence on the performance of AL methods, to the detriment of generalisability to other data sources. 

Active learning assumes the presence of a user interface to perform annotations but is only concerned with which data to annotate. Refinement assumes we can generate an annotation through iterative interaction with the current model prediction. Hence, it would be desirable to combine these two in future work. If we can train a model with a tiny amount of training data, and then ask annotators to refine model predictions towards a more accurate label, we can expedite the annotation process by reducing the initial annotation workload and reduce additional interface work for use with unseen data. This would be the same interface used to create the training annotations. By combining the efforts of active learning and iterative refinement into a unified framework we can rapidly produce annotations to train our model, as well as acquiring high quality results from our models from the beginning.  
This should also have the added side effect of training the model on data from the same distribution that it will be predicting on, reducing domain shift effects in unseen distributions. 

By incorporating our end user at each stage of the model life cycle we could also use human feedback on model performance to add a more 'human interpretable' metric of model confidence as each user could rank the performance of the model for each input as it sees it, potentially giving a metric of confidence based on human interpretation of the model output. This of course requires experts to be using the system.
One might argue that the models initial predictions may impart some influence over the human user but by crowd-sourcing the initial annotations to a less expert multi-label crowd we could reduce this bias.

Developments in uncertainty quantification will benefit both AL selection heuristics and interpretation of model outputs, but there is no guarantee that the best performing uncertainty metrics for selecting new samples to be annotated will be the same metrics that are the most interpretable to a human user.

\begin{figure*}[htb]
    \centering
    \includegraphics[height=0.8\paperheight]{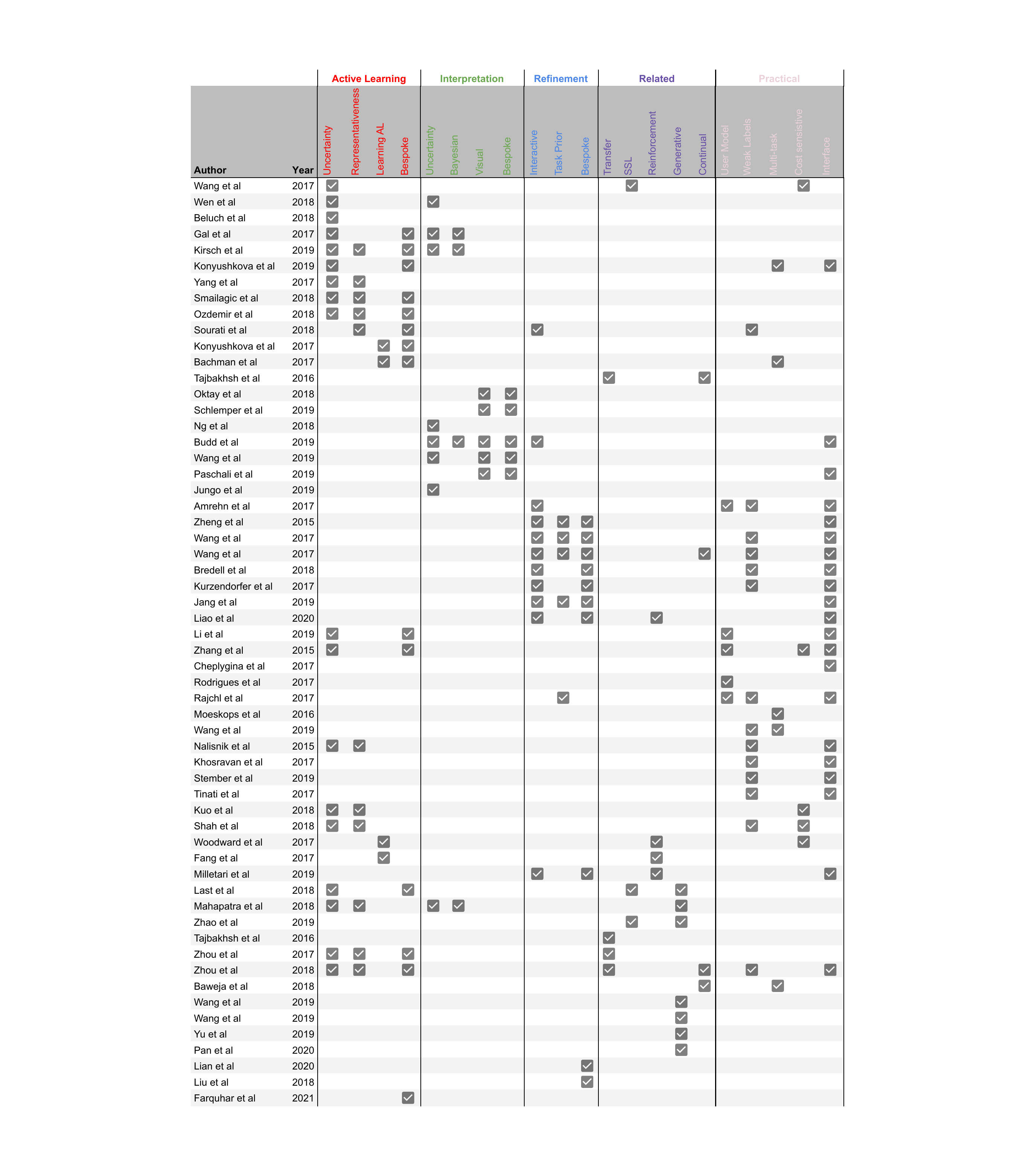}
    \caption{Table of features demonstrated by work discussed in this review}
    \label{fig:feature_table}
    \title{Feature Table}
\end{figure*}

Figure~\ref{fig:feature_table} outlines the core methods being used in human-in-the-loop computing for each of the papers discussed in this review. This figure shows that there is significant overlap of research goals for many areas of human-in-the-loop computing but there are large gaps that need to be filled in order to understand the relationships between different methods and how these might affect their performance.

As the many areas of DL research converge towards shared goals of working with limited training data to achieve state-of-the-art results, we expect to see more systems emerge that exploit the advances made in the range of sub-fields of ML described here. We have already seen the combination of several methods into individual frameworks but as of yet no works combine all of the approaches discussed into a single framework. As different combinations of approaches begin to appear it is important to consider the measure by which we assess their performance, as isolating individual developments becomes more difficult. Developing baseline human-in-the-loop methods to compare to will be vital to assess the contributions of individual works in each area and to better understand the influences of competing improvements in these areas.







\section{Conclusions}\label{sec:conc}
In this review we have explored the large body of emerging medical image analysis work in which a human end user is at the centre. Deep Learning has all the ingredients to induce a paradigm shift in our approach to a plethora of clinical tasks. The direct involvement of humans is set to play a core role in this shift. The works presented in this review each offer their own approaches to including humans in the loop and we suggest that there is sufficient overlap in many methods for them to be considered under the same title of Human-in-the-Loop computing. We hope to see new methodologies emerge that combine the strengths of AL and HITL computing into end-to-end systems for the development of deep learning applications that can be used in clinical practice. While there are some practical limitations as discussed, there are many proposed solutions to such issues and as research in these directions continues it is only a matter of time before deep learning applications blossom into fully-fledged, accurate and robust systems to be used for daily routine tasks. We are in an exciting era for medical image analysis, with endless opportunity to innovate and improve the current state-of-the-art and to leverage the powers of deep learning to make a real impact in health care across the board. With diligent research and development we should see more and more applications boosted by deep learning capabilities finding their way onto the market, allowing users to achieve better results, faster, and with less expertise than before, freeing up expert time to be used on the most challenging cases. The field of Human-in-the-loop computing will play a crucial role to achieve this.

\section*{Acknowledgments}
SB is supported by the EPSRC Centre for Doctoral Training in Smart Medical Imaging EP/S022104/1. This work was in part supported by EP/S013687/1, Intel and Nvidia. We thank 
Innovate UK: London Medical Imaging \& Artificial Intelligence Centre for Value-Based Healthcare [104691] for co-funding this research. 





\bibliographystyle{model2-names}\biboptions{authoryear}
\bibliography{references,extrefs}

\clearpage
\section*{Supplementary Material}

\end{document}